\documentclass[letterpaper, 10 pt, conference]{ieeeconf}  
\usepackage{multirow}
\usepackage{algorithm}
\usepackage{algorithmic}
\usepackage{bm}
\usepackage{subfigure}
\usepackage{graphicx}
\usepackage{amssymb}
\usepackage{cite}

\IEEEoverridecommandlockouts                              

\title{\LARGE \bf
	Multi-Scale Cost Volumes Cascade Network for Stereo Matching
}

\author{Xiaogang Jia, Wei Chen$^{*}$, Zhengfa Liang, Mingfei Wu, Yusong Tan, Libo Huang
	\thanks{$^{*}$Corresponding Author}%
	\thanks{Department of  Computer Science, National University of Defense Technology, China}%
	\thanks{{\tt\small  chenwei@nudt.edu.cn}}
}

\begin{document}

	\maketitle
	\thispagestyle{empty} 
	
	

	\begin{abstract}
		
		Stereo matching is essential for robot navigation. However, the accuracy of current widely used traditional methods is low, while methods based on CNN need expensive computational cost and running time. This is because different cost volumes play a crucial role in balancing speed and accuracy. Thus we propose MSCVNet, which combines traditional methods and neural networks to improve the quality of cost volume. Concretely, our network first generates multiple 3D cost volumes with different resolutions and then uses 2D convolutions to construct a novel cascade hourglass network for cost aggregation. Meanwhile, we design an algorithm to distinguish and calculate the loss for discontinuous areas of disparity result. According to the KITTI official website, our network is much faster than most top-performing methods (24×than CSPN, 44×than GANet, etc.). Meanwhile, compared to traditional methods (SPS-St, SGM) and other real-time stereo matching networks (Fast DS-CS, DispNetC, and RTSNet, etc.), our network achieves a big improvement in accuracy, demonstrating the feasibility and capability of the proposed method.
		
	\end{abstract}

	\section{Introduction}
	
	Stereo matching can recover the depth information of the scene from planar pictures, which is essential for depth estimation\cite{martins2018fusion,zhou2020padenet}, 3D reconstruction and person re-identification\cite{zeng2020hierarchical,zeng2020energy,ning2020deviation}, etc.  Since lidar is expensive, many robot applications rely on it to obtain depth for odometry and navigation. For example, the traditional SGM (Semi-Global Matching) algorithm\cite{hirschmuller2007stereo} is well known and widely used in real-time robot applications. However, the SGM algorithm is sensitive to uneven illumination and textureless areas \cite{xue2019multistereo}, which leads to low accuracy. As the profound research of Artificial Neural Network, Artificial Neural Network has been successfully applied in stereo matching. Thus most current methods adopt CNN (Convolutional Neural Network) to construct networks for stereo matching.
	
	According to the dimension of cost volumes and aggregation networks, stereo matching networks can be divided into the following two types: 2D convolution networks based on 3D cost volume\cite{yee2020fast,mayer2016large,xu2020aanet} and 3D convolution networks based on 4D cost volume\cite{kendall2017end,chang2018pyramid,cheng2019learning}. Recent top-performing methods mainly adopt the latter, such as GCNet\cite{kendall2017end}, PSMNet\cite{chang2018pyramid}, CSPN\cite{cheng2019learning}, etc. However, 3D convolution networks require a lot of memory and computational cost, so these methods often require many GPUs for training and long-running time for testing. Besides, complex network structures also lead to low generalization, far from replacing the SGM algorithm\cite{hirschmuller2007stereo} and applying it to real robot applications. 
	
	In contrast, 2D convolution networks\cite{mayer2016large,xu2020aanet} have significant advantages in memory and running time, but they lose a lot of feature information when generating 3D cost volume (all channel values of two pixels are converted into one value). This is very similar to traditional methods \cite{mei2011building} (the RGB channel values of two pixels are converted into one value). Therefore, both traditional methods and 2D convolution networks have low accuracy.
	
	Considering that a single 3D cost volume will lose a lot of feature information, and the 4D cost volume is far from being applied to real-time robot applications in terms of running time, we integrate traditional methods and CNN to generate multiple 3D cost volumes to replace the expensive 4D cost volume. At present, the running time of traditional methods and 2D convolution networks are both around 0.05s. Combining these two methods is still far less than 3D convolution networks, where the latter are mainly between 0.32s to 2s. At the same time, by combining the two methods, the accuracy will be greatly improved. Therefore, it is worthwhile to integrate traditional methods and 2D convolution networks.
	
	The followings are the detail contents and contributions of our method:
	
	\begin{itemize}
		\item We leverage multiple 3D cost volumes from the traditional method and CNN to improve stereo matching accuracy.
		\item We propose a fast and accurate stereo matching network, which is useful for real-time robot applications.
		\item We construct a novel 2D cascade hourglass network, which can effectively aggregate traditional methods and CNN features.
	\end{itemize}
	
	In the following sections, several closely related works are deeply discussed, and the details of our proposed method are fully expounded. Furthermore, we conduct intensive analyses and experiments to compare the difference in performance with other stereo networks. We also design ablation studies to evaluate the effectiveness of each module. All the results demonstrate the rationality and feasibility of our method.

	\section{Related work}
	
	Matching cost computation and cost aggregation are the fundamental and crucial steps of stereo matching. This section briefly introduces the overview of related works.

	\subsection{Traditional method for Matching cost computation}
	
	Traditional methods use the color, brightness, and gradient information of the images for stereo matching, including mutual information\cite{kim2003visual,egnal2000mutual}, Census transform\cite{ma2013modified,baik2006fast}, Rank transform\cite{banks1999constraint}, Birchfield and Tomasi\cite{birchfield1998pixel}, etc. These methods can generate a cost volume and rough disparity map quickly without any training process, but the accuracy is low. Take census transform as an example, census transform evaluates the similarity of different pixels by comparing the relationship between the center pixel with its surroundings, as shown in Equation~\ref{equ1}:
	\begin{equation}
	\label{equ1}
	\left\{
	\begin{array}{lr}
	C_{s}(u,v)=\mathop \otimes \limits_{i=-n'}^{n'} \mathop \otimes \limits_{j=-m'}^{m'} \zeta (I(u,v),I(u + i,v + j)) &  \\
	\zeta (x,y)=
	\left\{
	\begin{array}{lr}
	0 \qquad if \quad x \leq y &  \\
	1 \qquad if \quad x > y  &  
	\end{array}
	\right. &  
	\end{array}
	\right.
	\end{equation}
	where $n’$ and $m’$ are the largest integers not greater than half of $n$ and $m$ respectively, $\otimes$ is the bitwise concatenation operation. $I(u,v)$ denotes the pixel value at the location $(u,v)$. $C_{s}(u,v)$ denotes the census sequence at location $(u,v)$ for the image $s$.
	
	After calculating the census sequence of all pixels, traditional methods leverage the Hamming distance to measure the similarity between the reference pixel and candidate pixels, as shown in Equation~\ref{equ2}:
	\begin{equation}
	\label{equ2}
	C(d,u,v)=Hamming(C_{sl}(u,v),C_{sr}(u-d,v))
	\end{equation}
	
	However, there exists a great shortcoming for census transform\cite{ma2013modified,baik2006fast} to abandon the original image information. Researchers have to combine other traditional methods to supplement. Mei et al. propose the ADCensus algorithm\cite{mei2011building} through a combination of absolute difference (AD) and census transform. Recent Fast DS-CS\cite{yee2020fast} further extends the ADCensus method by integrating the classical neural network Unet\cite{ronneberger2015u}.

	\subsection{Neural network for Matching cost computation}

	Mayer et al. propose DispNetC\cite{mayer2016large}, a representative matching cost computation method, and use the 1D Correlation layer to generate the cost volume. Concretely, after capturing the image feature vectors by a neural network, DispNetC uses a dot product style operation to decimate the feature dimension and generate a cost volume of dimensionality H*W*D (D is the disparity candidate range). The 1D Correlation layer is defined as:
	\begin{equation}
	\label{equ3}
	C(d,x,y) = \frac {1}{N} \left \langle f_{l} \left( x,y \right),  f_{r} \left(x,y-d \right) \right \rangle
	\end{equation}
	where $f(x, y)$ is the feature vector at location $(x, y)$. $\left \langle \right \rangle$ is the inner product of two feature vectors, and $N$ denotes the channel number. 
	
	Another representative method GCNet\cite{kendall2017end} takes a different approach by concatenating the left feature with the corresponding right feature cross disparity range, and packing these into a 4D volume, which is defined as:
	\begin{equation}
	C(d,x,y) = Concat   \left\{ f_{l} \left( x,y \right),  f_{r} \left(x,y-d \right) \right\}
	\end{equation}
	
	However, 4D cost volumes need many 3D convolutions for cost aggregation. These networks\cite{zhang2019ga,cheng2019learning,kendall2017end} have expensive computational cost and running time, which is far from meeting the need for real-time applications.
	
	\begin{figure*}[]
		\centering
		\includegraphics[width=.95\linewidth]{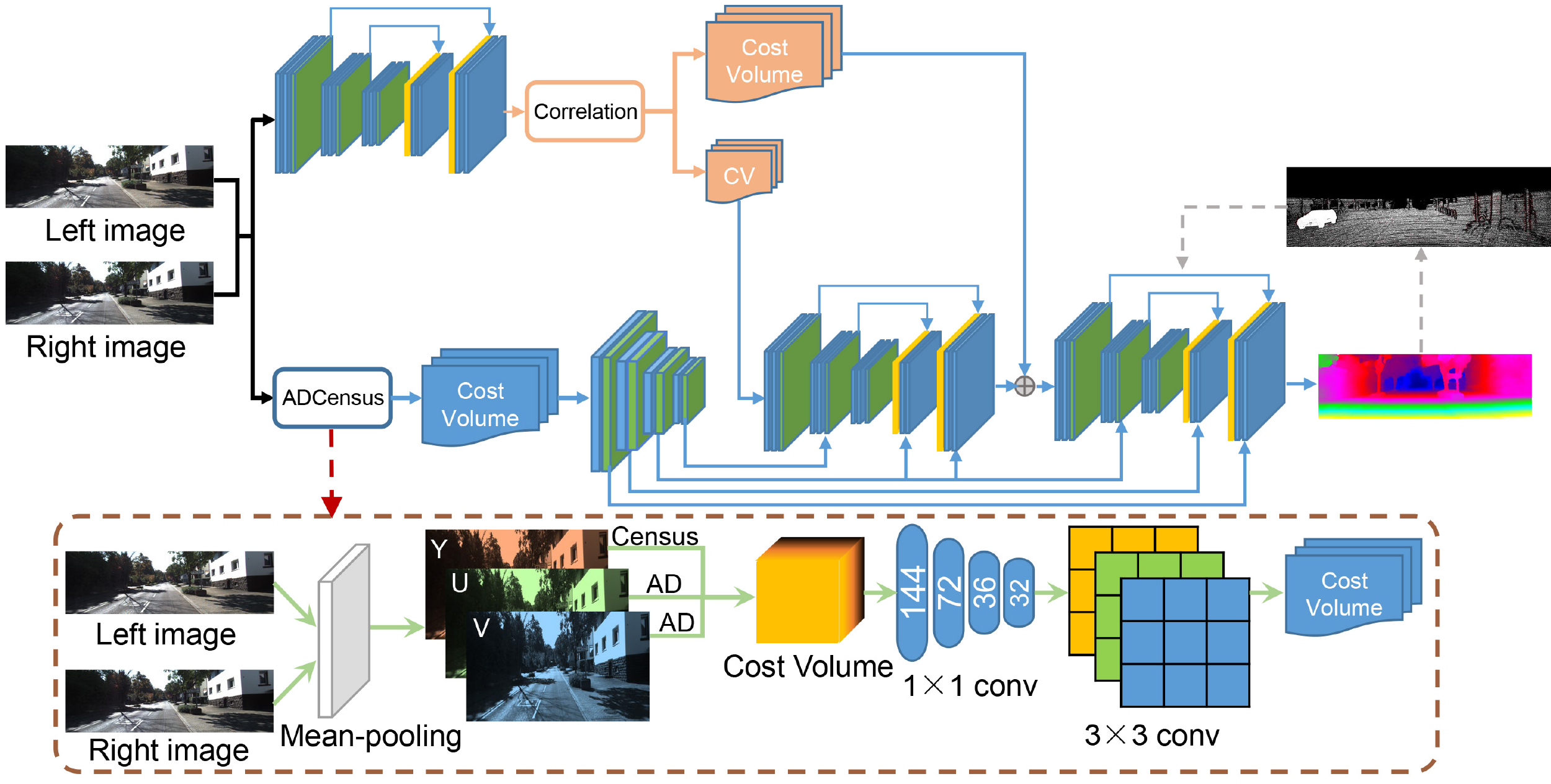}
		\caption{\label{fig1}
			Architecture overview of proposed MSCVNet. We use census transform and Absolute difference to construct the initial cost volumes and then use per-pixel convolutions to adjust the cost volumes to a low-dimensional feature vector. Meanwhile, after extracting image features through a small Unet, we use a 1D Correlation layer to generate two cost volumes with different dimensions. The traditional cost volume is leveraged by an encoder to predict multi-scale guidance features, which will be fed into two cascade hourglass networks with other cost volumes to predict the final disparity map.}
	\end{figure*}

	\subsection{Multi-Scale Cost Volumes}
	
	Most current stereo matching networks generate one cost volume in matching cost computation. However, 3D cost volumes lose much feature information, while 4D cost volumes require many 3D convolutions, leading to a significant increase in GPU memory and running time. Therefore, it isn't easy to a trade-off between accuracy and speed by using only a single cost volume. Several works adopt a coarse-to-fine architecture that leverages multi-scale cost volumes to improve accuracy to mitigate this problem. Yang et al.\cite{yang2019hierarchical} design a hierarchical coarse-to-fine network that builds up a pyramid of cost volumes. Gu et al.\cite{gu2020cascade} introduce an efficient cost volume calculation method with several cost volumes at different resolutions and corresponding disparity intervals. Similarly, Xu et al.\cite{xu2020aanet} propose AANet, which constructs multiple 3D cost volumes by correlating image features at different scales and then aggregates them with several stacked Adaptive Aggregation Modules. All these methods demonstrate that multi-scale cost volumes can achieve competitive accuracy while maintaining fast speed.
	
	\section{Method}
	
	We present MSCVNet, which mainly contains the following two modules: Multi-scale 3D cost volumes and Guided cascade hourglass network. Unlike previous multi-scale cost volume methods, our network combines traditional methods and CNN to generate multi-scale cost volumes with different resolutions, thus avoiding feature redundancy caused by a single method, which will be discussed in the ablation study. Meanwhile, our guided cascade hourglass structure can fully extract features and effectively integrate the traditional method and CNN to improve accuracy. The architecture of MSCVNet is illustrated in Fig.~\ref{fig1}.
	
	
	\subsection{Multi-scale 3D cost volumes}
	
	\subsubsection{Traditional Method}
	We first introduce the process of generating cost volume from the traditional method. To reduce the computational cost, we use mean-pooling to adjust the stereo images to 1/2 resolution. Then the images are converted from RGB to YUV. For the Y channel, we use the 5*5 census transform to compute the census sequence of the stereo images. Then we calculate the hamming distance across the maximum disparity range in the same horizontal direction, as mentioned in Equation~\ref{equ2}. As a rule of thumb,  the maximum disparity candidate range is set to 192. Therefore, we can get the census cost volume $C_{1}$ with size $1/2H*1/2W*96$. Since the census transform can't match occlusion and textureless areas well and the absolute difference method is sensitive to the pixels' difference, we use AD on the U and V channels to construct the other two cost volumes $C_{2}$ and $C_{3}$. After getting all the three cost volumes, we concatenate the features into one 3D tensor of $1/2H*1/2W*288$. For each pixel $(x, y)$, the cost can be expressed as follows:
	
	$C(x,y)=[C_{1}(x,y,0),C_{2}(x,y,0),...,C_{3}(x,y,95)]$
	
	To speed up the network training, we normalize the above cost volume to satisfy the distribution with zero mean and unit variance. Afterward, we use four 1*1 convolution layers to reduce dimensions to 144,72,36,32. Furthermore, we concatenate the left input image to the succinct low-dimensional cost volume and use three 3*3 convolution layers for feature harvesting. Finally, we can get the traditional cost volume with a size of $1/2H*1/2W*32$.
	
	\subsubsection{Neural Network}
	
	We use 1D Correlation to construct the other two cost volumes. First, we use a feature extractor to capture multi-scale features. Our feature extractor is an Unet structure\cite{ronneberger2015u}, an encoder-decoder with skip connections and learnable parameters. More concretely, we implement downsampling by a 3*3 convolution followed by a 2*2 convolution with the stride of 2. We achieve the up-sampling block by 2*2 deconvolution with the stride of 2 and use skip-connection to concatenate features with the same resolution. We leverage the 1*1 convolution to reduce the dimension by half, followed by a 3*3 convolution for feature harvesting. Each convolution layer uses batch normalization and Relu for non-linearities. We only use the last feature map of each resolution for correlation and further upsample to generate a higher resolution feature map. Thus the high-resolution feature map contains part of spatial context information. Finally, we can obtain multi-scale feature maps of the left and right images.
	
	We only use feature maps with 1/2 and 1/4 resolution for correlation. Their max disparity ranges are 96 and 48, respectively. Therefore, we can obtain the other two cost volumes with a size of $1/2H*1/2W*96$ and $1/4H*1/4W*48$. Similar to the traditional method, we use 1*1 convolution to convert the 1/2 resolution cost volume into a low-dimensional vector of $1/2H*1/2W*32$.
	
	\subsection{Guided cascade hourglass network}
	
	Our cost aggregation network mainly consists of the following two modules: Multi-scale guided feature encoder and Cascade hourglass network. The specific implementation of our network is as follows:
	
	\begin{figure}[t]
		\centering
		\includegraphics[width=.95\linewidth]{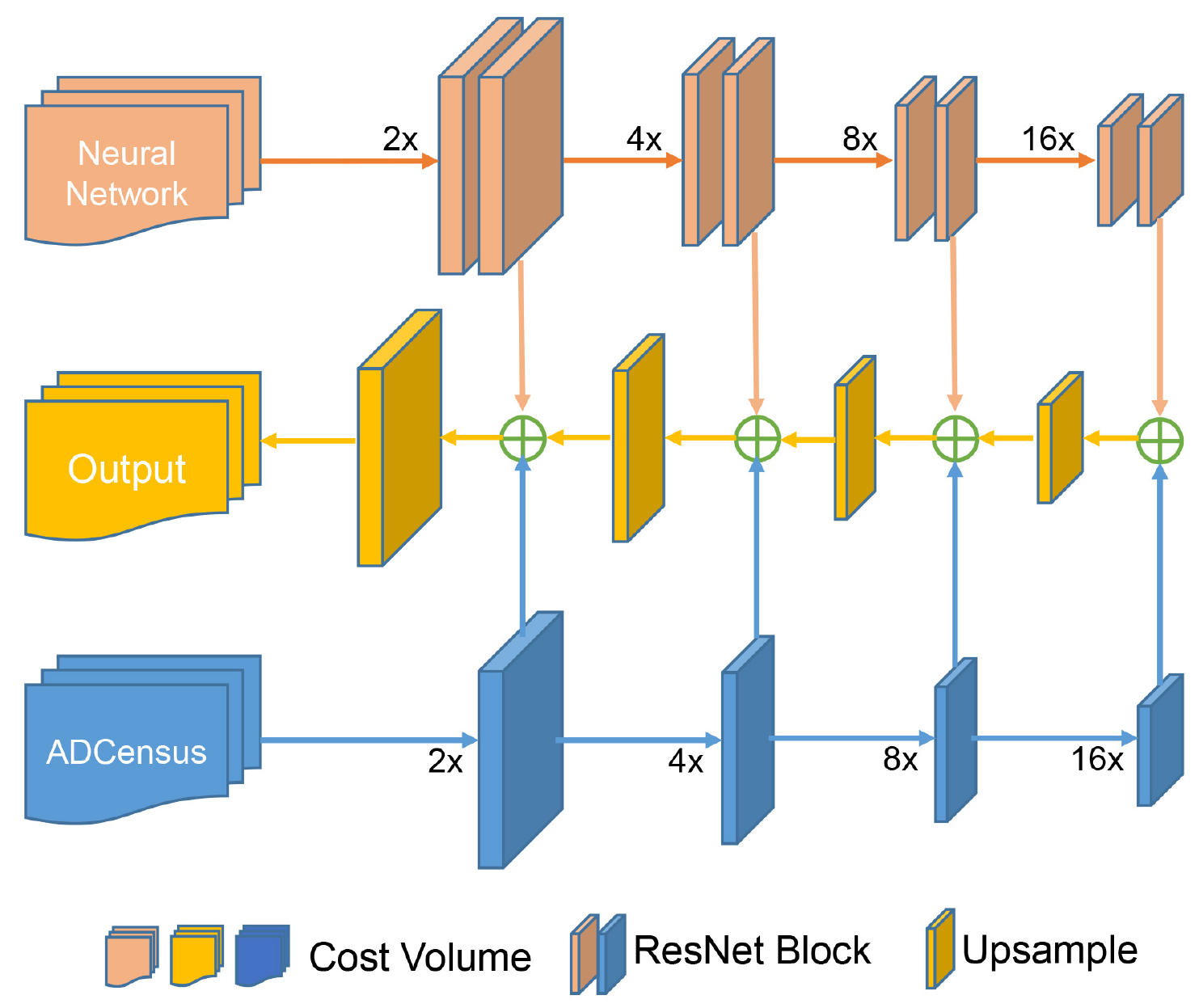}
		\caption{\label{fig2}
			The detailed structure of the cascade hourglass network. We use residual blocks to extract features with multiple resolutions from CNN and combine them with the traditional method in a decoder.}
	\end{figure}
	
	\subsubsection{Multi-scale Guided Feature Encoder}
	
	As shown in Fig.~\ref{fig1}, we first capture multi-scale guidance features based on the traditional cost volume. Our down-sampling block applies two 3*3 convolutions with the stride of 2 and 1 separately. The deepest feature map reaches a 1/16 spatial resolution after three layers of down-sampling convolutions. Therefore, the traditional cost volume can provide four different scale guide features for the subsequent cascade network.
	
	\begin{figure}
		\centering
		\subfigure[Input Left Image]{
			\includegraphics[width=0.45\textwidth]{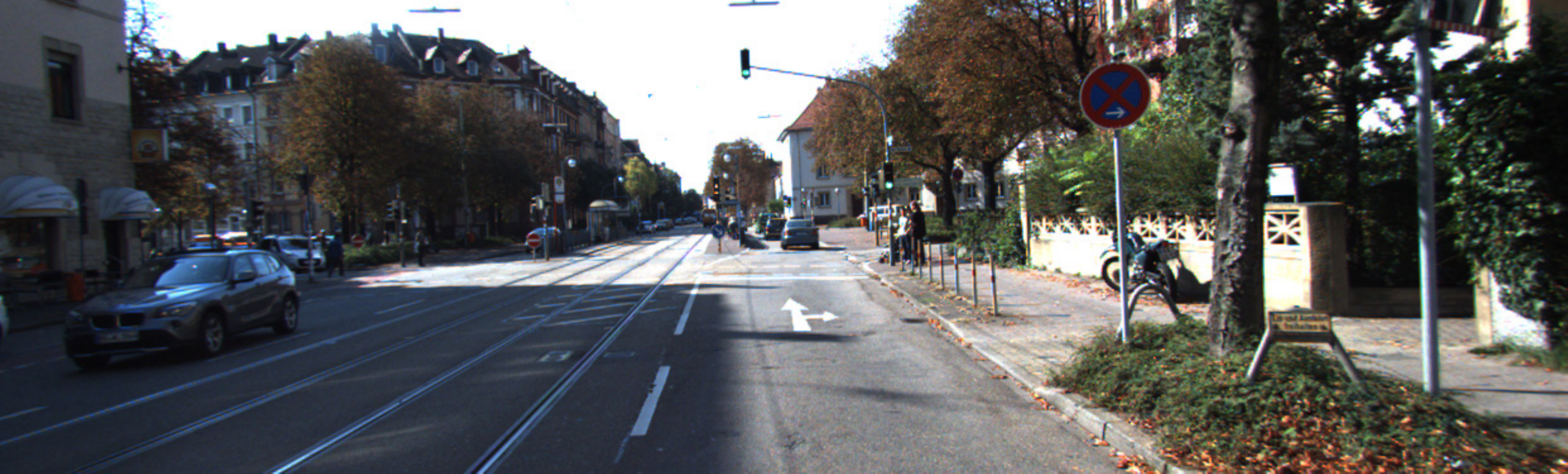} 
		}
		\subfigure[Warped Left Image]{
			\includegraphics[width=0.45\textwidth]{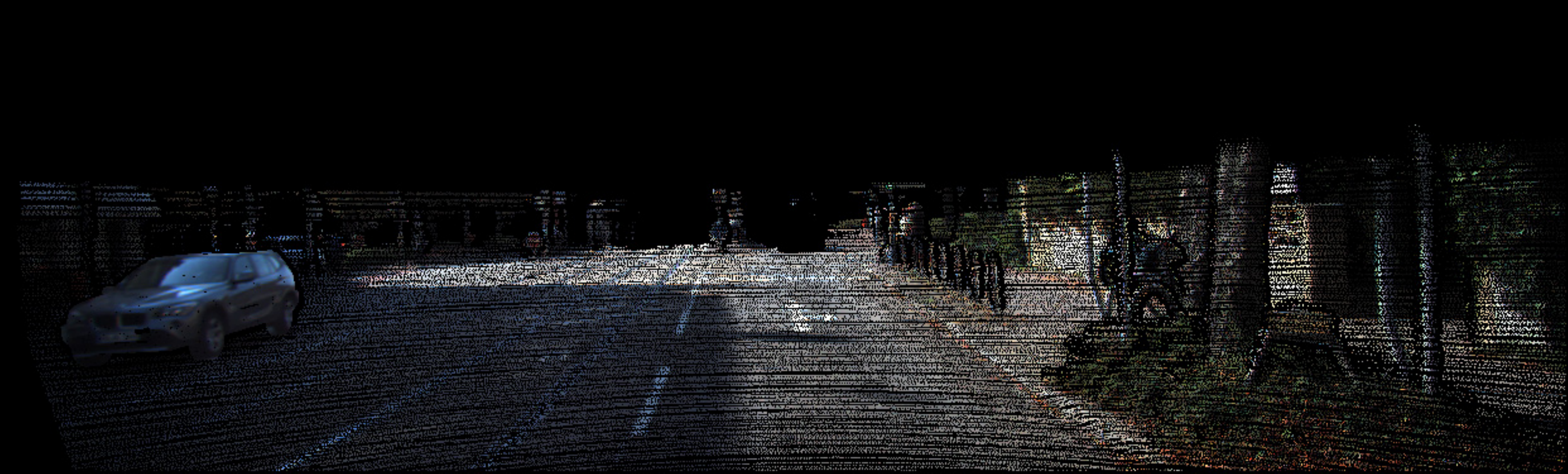} 
		}
		\subfigure[Warped Disparity Image and Discontinuous Areas]{
			\includegraphics[width=0.45\textwidth]{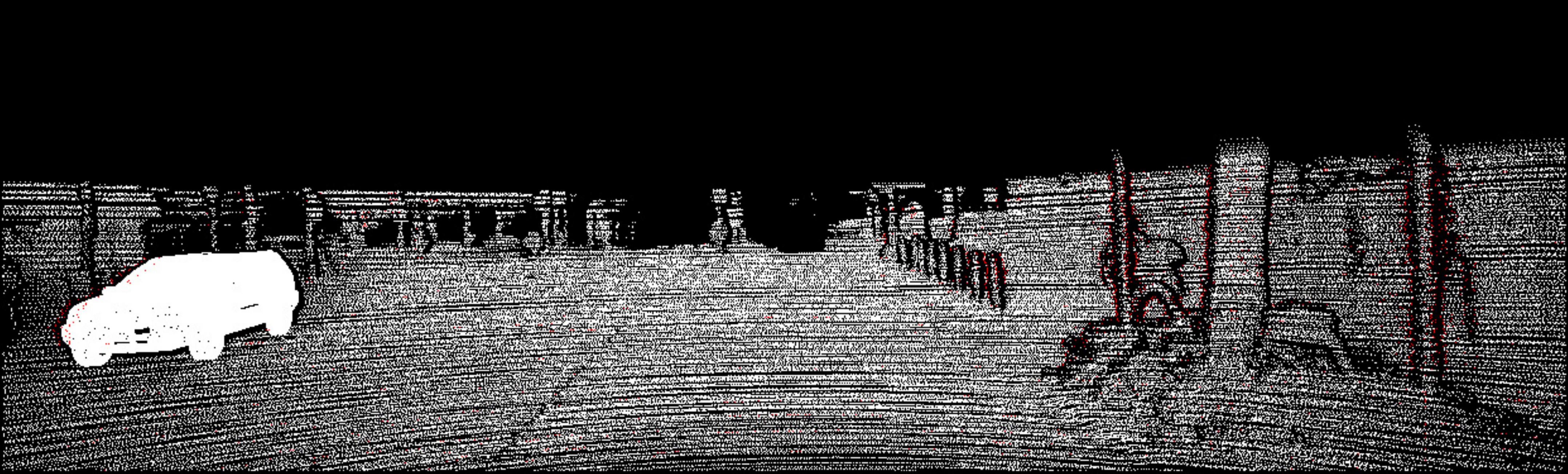} 
		}
		\caption{\label{fig3}
			Left image of different domains. We calculate (b) based on the left disparity image and the right image. As we can see, there are artifacts near some objects, such as trees, telephone poles, and cars. The artifacts make the warped disparity map no longer increase monotonically. Therefore, we extract discontinuous disparity areas in (c).}
	\end{figure}
	
	\subsubsection{Cascade Hourglass Network}
	
	We use an encoder to extract feature vectors from the other two cost volumes, leveraged by a decoder to cooperate with the features generated from traditional cost volume. The architecture of one cascade hourglass network is shown in Fig.~\ref{fig2}.

	The first hourglass network uses a series of ResNet\cite{he2016deep} blocks with a stride of 2 to downsize the feature resolution to 1/16 gradually. Then the decoder consists of three up-sampling blocks to increase the feature resolutions progressively and integrate guided features. Since the resolution of the output is 1/4, we use a deconvolution layer with a stride of 2 to up-sample the output and then combine the result with another cost volume as the following hourglass network's input. The last hourglass network focuses more on generating high-resolution features, while the first hourglass network focuses on extracting deep features.
	
	After the two cascade hourglass networks, we can obtain a refined cost volume with size $1/2H*1/2W*32$. Then we use a 1*1 convolution to generate the disparity map directly and use bilinear interpolation to upsample this map to the full resolution.
	
	\begin{table*}[]
		\caption{\label{tab1}Performance comparison on the Scene Flow and KITTI. We select several classical accurate or fast networks for comparison. Our MSCVNet can achieve competitive performance with significantly fast speed. }
		\centering
		\begin{tabular}{l|c|cc|cc|ccc|ccc|r}
			\hline
			\multirow{3}{*}{Method} & Scene Flow           & \multicolumn{4}{c|}{KITTI2012}                                                  & \multicolumn{6}{c|}{KITTI2015}                                             & \multirow{3}{*}{Time(s)} \\ \cline{2-12}
			& \multirow{2}{*}{EPE} & \multicolumn{2}{c|}{\textgreater{}3px} & \multicolumn{2}{c|}{\textgreater{}5px} & \multicolumn{3}{c|}{All pixels} & \multicolumn{3}{c|}{Non-Occluded pixels} &                          \\
			&                      & \textbf{Out-Noc}       & Out-All       & \textbf{Out-Noc}       & Out-All       & D1-bg & D1-fg & \textbf{D1-all} & D1-bg    & D1-fg    & \textbf{D1-all}    &                          \\ \hline
			MC-CNN\cite{vzbontar2016stereo}                  & 3.79                 & 2.43                   & 3.63          & 1.64                   & 2.39          & 2.89  & 8.88  & 3.89            & 2.48     & 7.64     & 3.33               & 67                       \\
			GC-Net\cite{kendall2017end}                 & 2.51                 & 1.77                   & 2.30          & 1.12                   & 1.46          & 2.21  & 6.16  & 2.87            & 2.02     & 5.58     & 2.61               & 0.9                      \\
			PSMNet\cite{chang2018pyramid}                  & 1.09                 & 1.49                   & 1.89          & 0.90                   & 1.15          & 1.86  & 4.62  & 2.32            & 1.71     & 4.31     & 2.14               & 0.41                     \\
			GANet\cite{zhang2019ga}                   & 0.84                 & 1.19                   & 1.60          & 0.76                   & 1.02          & 1.48  & 3.16  & 1.81            & 1.34     & 3.11     & 1.63               & 1.8                      \\ \hline
			SPS-St\cite{yamaguchi2014efficient}                  & -                    & 3.39                   & 4.41          & 2.33                   & 3.00          & 3.84  & 12.67 & 5.31            & 3.5      & 11.61    & 4.84               & 2                        \\
			RTSNet\cite{lee2019real}                  & -                    & 2.43                   & 2.90          & 1.42                   & 1.72          & 2.86  & 6.19  & 3.14            & 2.67     & 5.83     & 3.19               & 0.02                     \\
			Fast DS-CS\cite{yee2020fast}              & -                    & 2.61                   & 3.20          & 1.46                   & 1.85          & 2.83  & 4.31  & 3.08            & 2.53     & 3.74     & 2.73               & 0.02                     \\
			BSDCNet\cite{jia2021bidirectional}              & -                    & 2.39                   & 2.92          & 1.35                   & 1.68          & 2.49  & 4.98  & 2.90            & 2.27     & 4.48     & 2.64               & 0.025                     \\
			DispNetC\cite{mayer2016large}                & 1.68                 & 4.11                   & 4.65          & 2.05                   & 2.39          & 4.32  & 4.41  & 4.34            & 4.11     & 3.72     & 4.05               & 0.06                     \\ \hline
			MSCVNet                 & 1.32                 & 2.25                   & 2.81          & 1.37                   & 1.74          & 2.31  & 5.41  & 2.82            & 2.12     & 5.02     & 2.60               & 0.041                    \\ \hline
		\end{tabular}
	\end{table*}
	
	\begin{figure*}
		\centering
		\subfigure[Input Stereo Pair]{
			\begin{minipage}[b]{0.23\textwidth} 
				\includegraphics[width=1\textwidth]{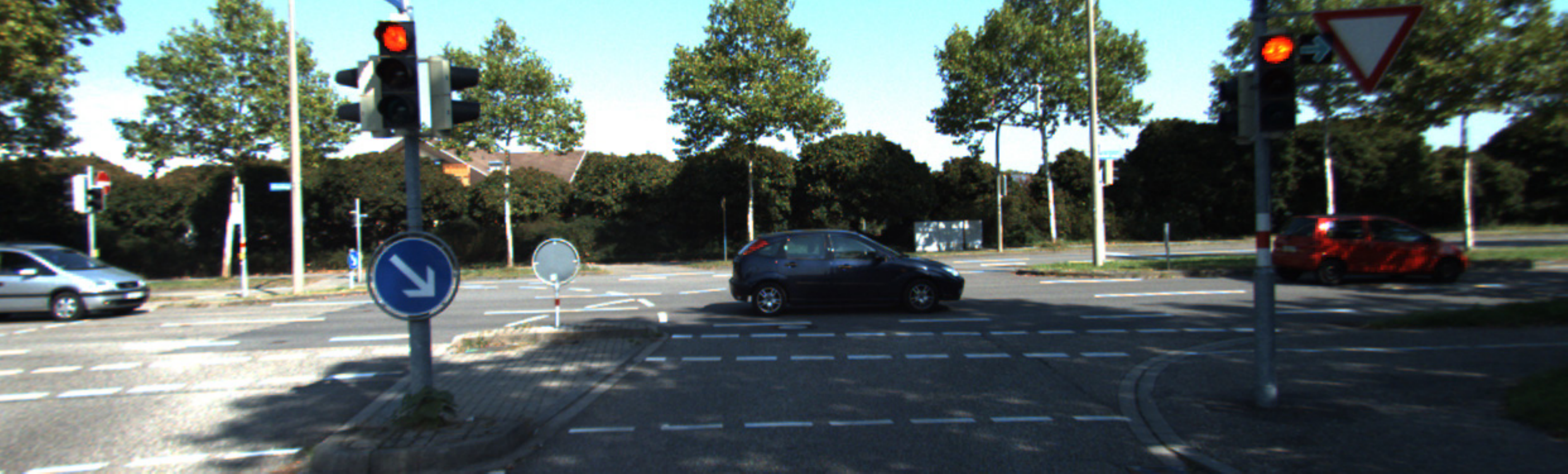} \\ 
				
				\includegraphics[width=1\textwidth]{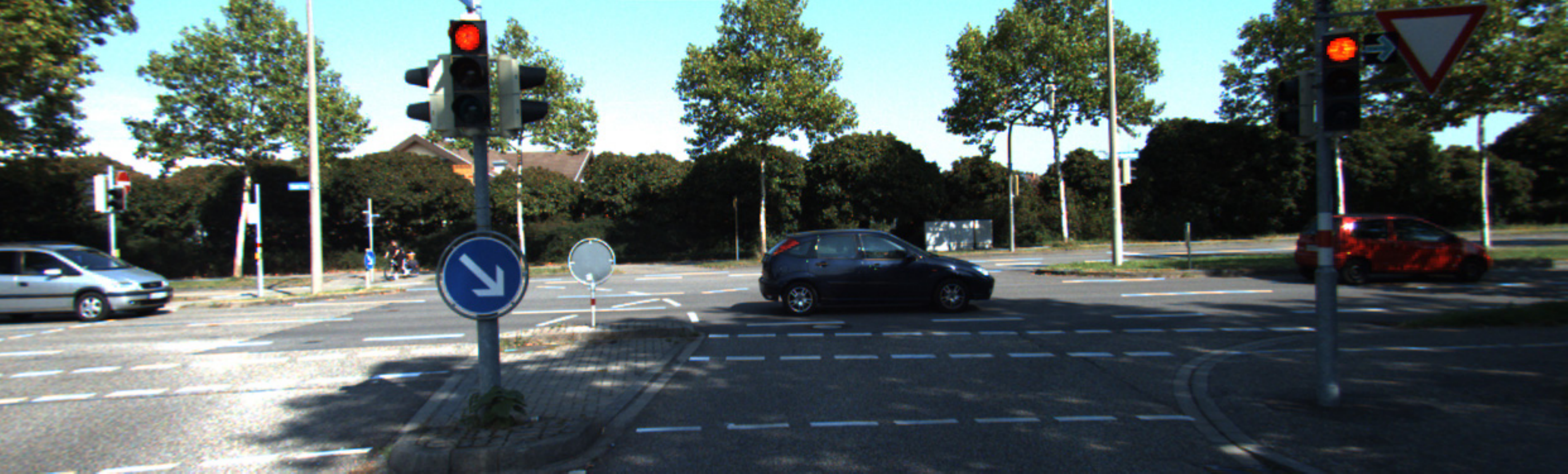}
			\end{minipage}
		}
		\subfigure[MSCVNet]{
			\begin{minipage}[b]{0.23\textwidth} 
				\includegraphics[width=1\textwidth]{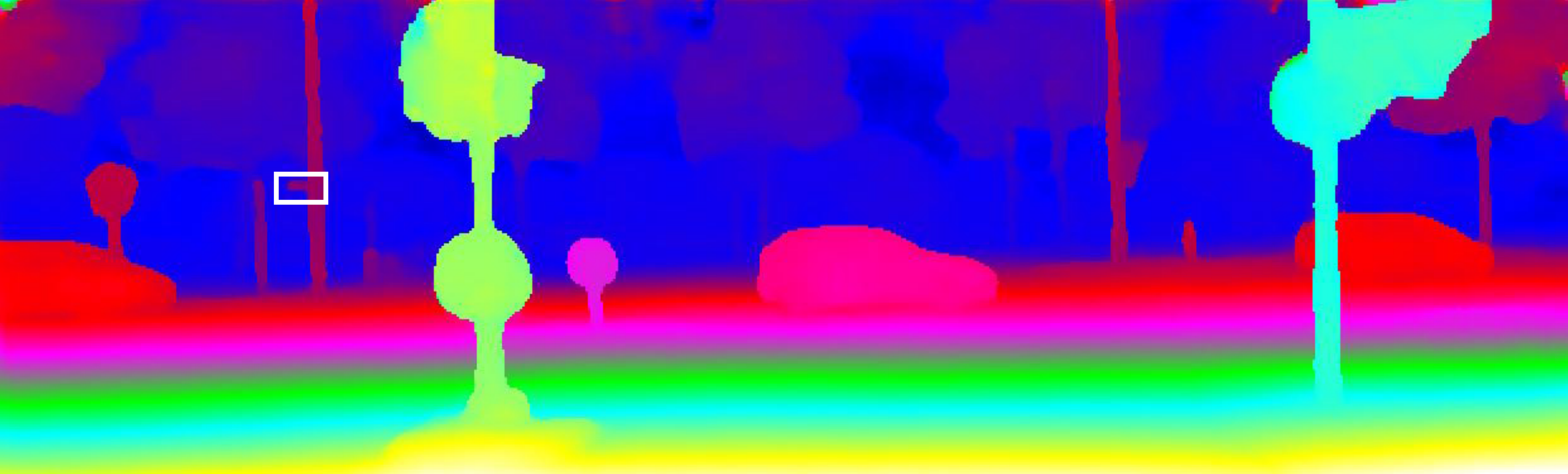} \\ 
				
				\includegraphics[width=1\textwidth]{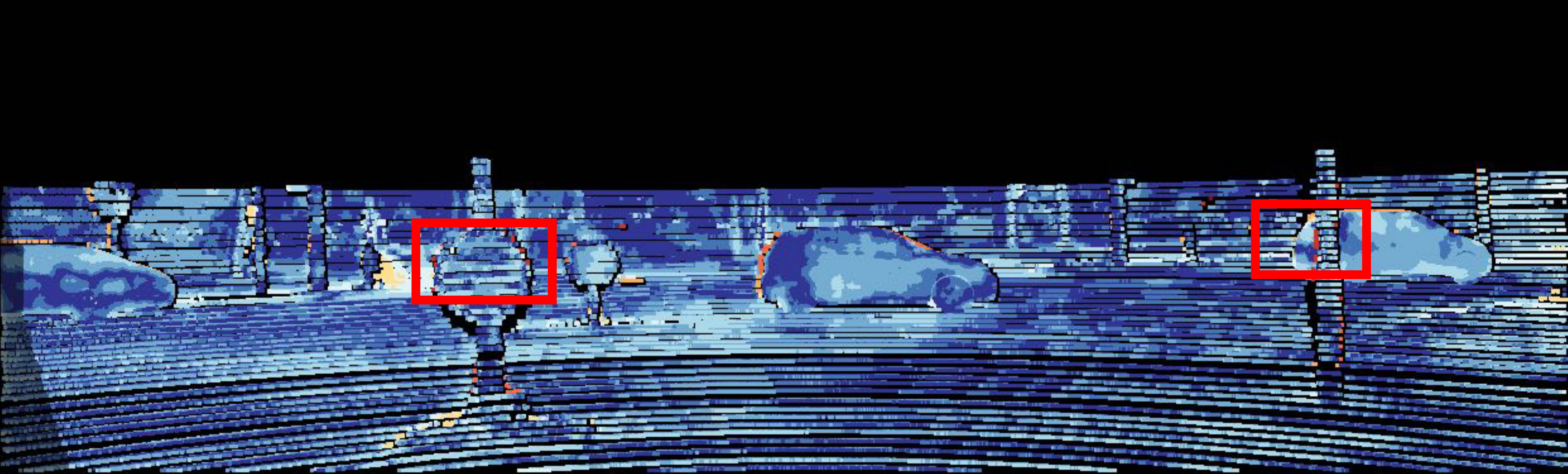} 
			\end{minipage}
		}
		\subfigure[Fast DS-CS]{
			\begin{minipage}[b]{0.23\textwidth} 
				\includegraphics[width=1\textwidth]{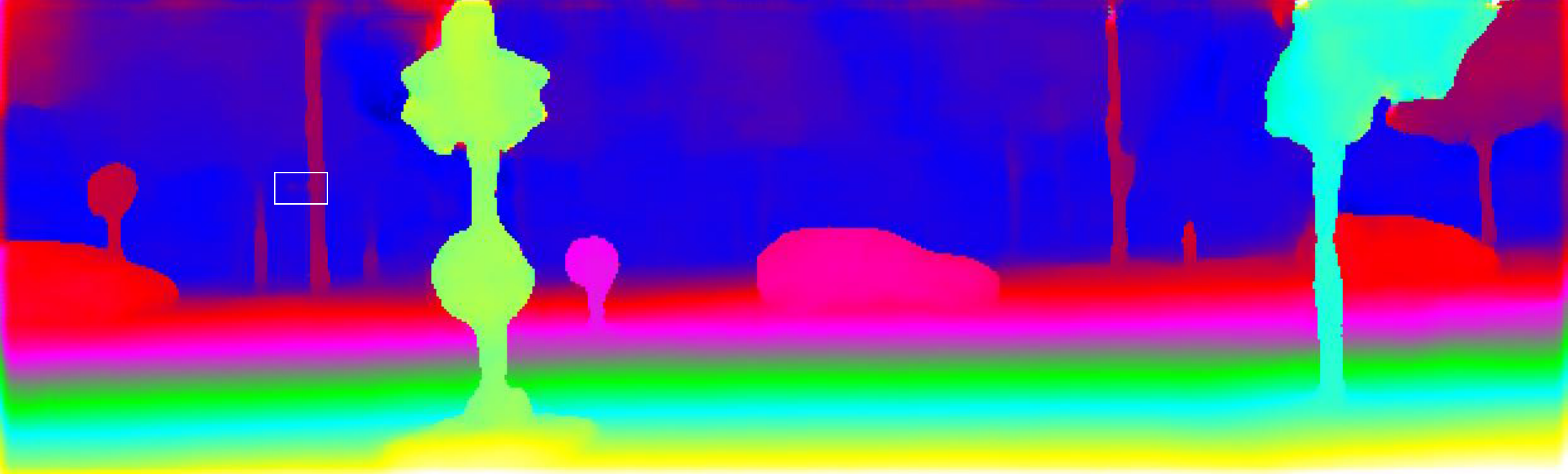} \\ 
				
				\includegraphics[width=1\textwidth]{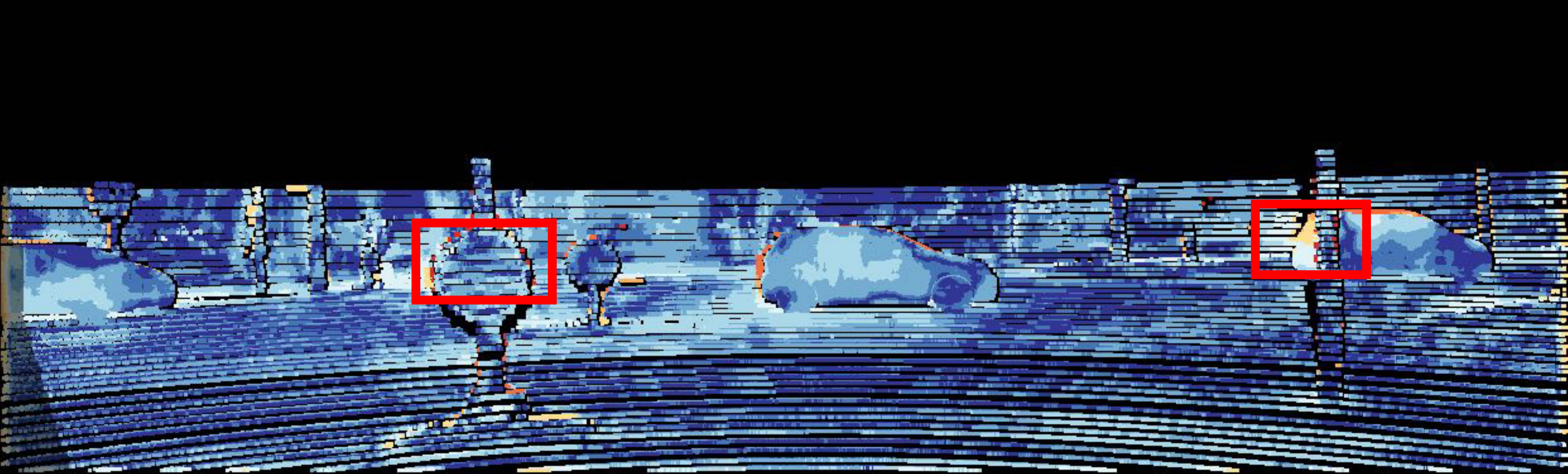} 
			\end{minipage}
		}
		\subfigure[DispNetC]{
			\begin{minipage}[b]{0.23\textwidth} 
				\includegraphics[width=1\textwidth]{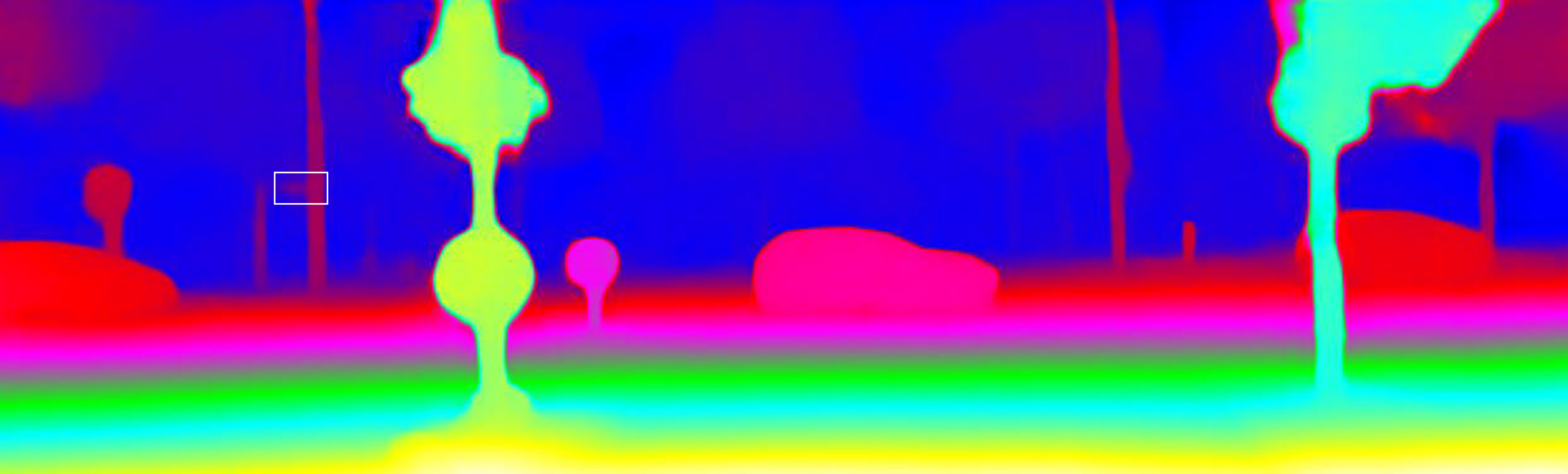} \\ 
				
				\includegraphics[width=1\textwidth]{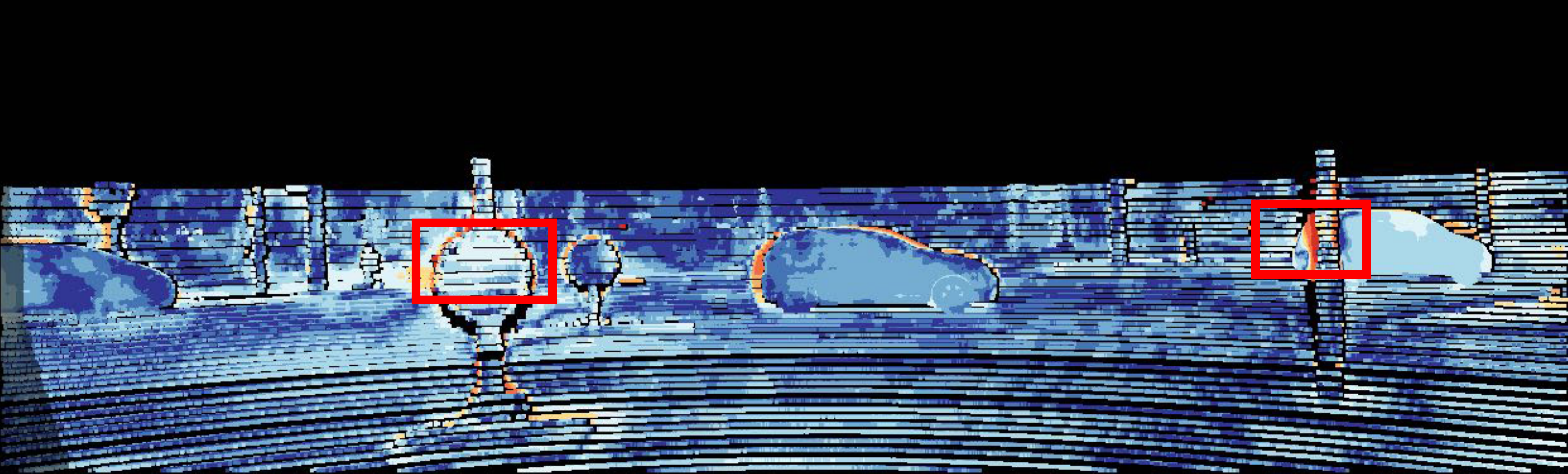} 
			\end{minipage}
		}
		\centering
		\caption{\label{fig4}
			Results of KITTI 2015 official website. Our MSCVNet can recover more details and alleviate the well-known edge-fattening issue. All disparity and error maps are collected from the KITTI2015 official website.}
	\end{figure*}
	
	\subsection{Training Loss}
	
	We propose a novel algorithm, which divides the disparity map into two parts: monotonically increasing areas (including occlusion), and discontinuous areas, as shown in Fig.~\ref{fig3}. Thus our network can improve the accuracy of different areas by changing the weights. 
	
	Our algorithm uses the following steps to calculate the discontinuous areas of any disparity map. We first generate warped disparity image $Warp(D)$ according to the left disparity map and coordinate value. Due to occlusion areas, $Warp(D)$ isn't monotonically increasing, and discontinuous areas will appear at the edge of some objects. Simply put, we analyze a row of pixels in $Warp(D)$, which can be extended to the entire disparity map.
	
	First, we assume that $Y$ is a row of pixels in $Warp(D)$, which can be expressed as follows:
	
	$Y=\left\{Y_{1},Y_{2},...,Y_{m},y_{1},y_{2},...,y_{n},Y_{k},Y_{k+1},...\right\}$
	
	where $Y_{1}<Y_{2}<...<Y_{m}<Y_{k}$, $y_{1}<y_{2}<...<y_{n}<Y_{m}$. As shown in Fig.~\ref{fig3}, for large objects, such as trees, cars, houses, etc. the disparity of these areas will suddenly decrease rather than suddenly increase. Their discontinuous disparity areas are $Y_{m}$, $y_{1}$. For small objects, such as railings, grass and telephone poles, etc. the disparity of these areas suddenly decreases and increases. Their discontinuous disparity areas are $Y_{m}$, $y_{1}$ and $y_{n}$, $Y_{k}$. We distinguish these two cases by comparing the value of $Y_{k}-y_{n}$ and $\epsilon$.
	
	The output of our algorithm is:
	
	$Output=
	\left\{
	\begin{array}{lr}
	0,0,...,1,1,0,...,1,1,0,...   \quad Y_{k}-y_{n} \leq \epsilon &  \\
	0,0,...,1,1,0,...,0,0,0,...   \quad Y_{k}-y_{n} > \epsilon  &  
	\end{array}
	\right.$
	
	We first calculate the maximum pixel values of the current coordinate:
	
	$Y_{max}=\left\{Y_{1},Y_{2},...,Y_{m},Y_{m},Y_{m},...,Y_{m},Y_{k},Y_{k+1},...\right\}$
	
	Then we calculate the absolute difference with Y and analyze the pixels greater than 0 and $\epsilon$:
	
	$Mask(Y_{max}-Y)=\left\{0,0,...,0,1,1,...,1,0,0,...\right\}$
	
	After that, we move Mask to the left and right by one bit and calculate the absolute difference with the original Mask.
	
	$|Mask_{>>1}-Mask|=\left\{0,0,...,0,1,0,...,0,1,0,...\right\}$
	
	$|Mask_{<<1}-Mask|=\left\{0,0,...,1,0,0,...,1,0,0,...\right\}$
	
	Finally, we can get the $Output$ by adding the two results together. 
	
	Based on Fast DS-CS\cite{yee2020fast}, our loss function is defined as:
	\begin{equation}
	L(d_{GT},\hat{d}) = \max \left( \tau, \left(d_{GT}-\hat{d}\right)*\left(1-\lambda*Output \right)\right)^{1/8}
	\end{equation}
	where $\hat{d}$ and $d_{GT}$ denote the final disparity map from our network and the ground truth, respectively. $Output$ denotes disparity classification map. Since the maximum disparity range is 192, and we set $\tau$ to 1, all pixels' loss value is adjusted to the range of 1-1.93. Meanwhile, we set $\lambda$ to 0.5, which can reduce the impact on the loss of discontinuous disparity areas and improve accuracy. Since the test data's ground truth is sparse, we only compute the loss for valid pixels (disparity $<$ 192).

	\section{Experiments}
	
	We evaluate our MDCNet on three popular public datasets: Scene Flow\cite{mayer2016large}, KITTI 2012\cite{Geiger2012CVPR} and KITTI 2015\cite{Menze2015ISA}, which consists of comparisons with the latest stereo methods and extensive ablation studies. Since there are few training images in the real datasets, we first conduct pre-training on the synthetic dataset Scene Flow. Then we fine-tune on the real dataset KITTI 2012 and KITTI 2015. We select the best training model for testing and put the results on KITTI official website for comparison. Finally, we conduct several ablation studies using KITTI2015 to evaluate the impact on performance and running time with different cost volumes, the number of cascade hourglass networks, and different loss functions.
	
	\begin{table*}[]
		\caption{\label{tab2}Evaluation of network with different settings. We use the percentage of three-pixel-error of all pixels (PC3), the mean disparity error (EPE) and runtime as evaluation metrics.}
		\centering
		\begin{tabular}{|c|c|c|c|c|c|c|c|c|c|}
			\hline
			& \multicolumn{2}{c|}{Cost Volume} & \multicolumn{2}{c|}{Cost Aggregation} & \multicolumn{2}{c|}{Loss} & \multicolumn{3}{c|}{KITTI2015} \\ \hline
			& ADCensus        & 1D Correlation        & Hourglass*1       & Hourglass*2       & $\lambda$=0         & $\lambda$=0.5         & PC3(\%)   & EPE    & Time(s)   \\ \hline
			1 & \checkmark               &                & \checkmark                 &                   & \checkmark          &              & 97.34     & 0.78   & \textbf{0.021}     \\ \hline
			2 &                 & \checkmark              & \checkmark                 &                   & \checkmark          &              & 97.78     & 0.73   & 0.023     \\ \hline
			3 & \checkmark               & \checkmark              & \checkmark                 &                   & \checkmark          &              & 98.05     & 0.64   & 0.035     \\ \hline
			4 & \checkmark               & \checkmark              &                   & \checkmark                 & \checkmark          &              & 98.31     & 0.61   & 0.041     \\ \hline
			5 &                 & \checkmark              &                   & \checkmark                 & \checkmark          &              & 97.93     & 0.74   & 0.037     \\ \hline
			6 & \checkmark               & \checkmark              &                   & \checkmark                 &            & \checkmark            & \textbf{98.54}     & \textbf{0.59}   & 0.041     \\ \hline
		\end{tabular}
	\end{table*}

	\subsection{Network Details}
	
	Our network is based on Tensorflow and uses Adam as optimizer$(\beta _{1}=0.9, \beta _{2}=0.999)$. We use an NVIDIA GTX 1080Ti GPU for training with a batch size of 4. Since many convolutions in our network don't use the BN layer, we set the maximum learning rate to $10^{\textnormal{-}4}$. We begin by training 500k iterations on Scene Flow with the learning rate of $10^{\textnormal{-}4}$. For KITTI, we use the final model pre-trained on Scene Flow and then fine-tune for 100k iterations and another 50k with a learning rate of $10^{\textnormal{-}4}$ and $10^{\textnormal{-}5}$ respectively. We combine the KITTI2012 and KITTI2015 datasets and train them together. We train the final model for 50k with the learning rate of $2*10^{\textnormal{-}5}$ on KITTI2012 and KITTI2015 separately.
	
	We also conduct data augmentation in the following ways: For Scene Flow and KITTI2015, since these two datasets provide ground-truth of both left and right images, we can get additional training data by flipping the stereo images and the right disparity map horizontally. Meanwhile, we use random cropping and vertical flipping for the training data of KITTI and Scene Flow.
	
	\subsection{Results on Scene Flow}
	
	Scene Flow is a large-scale synthetic dataset consisting of 34540 stereo pairs for training and 2347 stereo pairs for testing. This dataset provides a large number of dense and accurate disparity maps as the ground truth. We use the end-point error (EPE) as the evaluation metric, which denotes the average pixel error. 
	
	After training for 500k iterations, we compare the performance with other latest stereo matching methods. The final results are shown in Table~\ref{tab1}. As shown in the table, our MSCVNet can achieve a competitive result at a fast speed. Since we only pre-train on the Scene Flow without fine-tuning, the performance can be further improved. We mainly conduct a detailed comparison and analysis on KITTI2012 and KITTI2015.
	
	\subsection{Results on KITTI}
	
	KITTI is a real-world dataset with street scenes from a driving car. This dataset provides sparse but accurate dense disparity maps as ground truth. Image size is H = 376 and W = 1240. For KITTI2012\cite{Geiger2012CVPR}, it consists of 200 stereo images with ground-truth disparities for training and 200 image pairs without ground-truth disparities for testing. For KITTI2015\cite{Menze2015ISA}, there are 194 stereo images for training and 195 for testing. During training, we combine KITTI2012 and KITTI2015 and divide the whole training images into a training set with 354 image pairs and a validation set with 40 image pairs (20 from KITTI2012 and 20 from KITTI2015). We evaluate our method using official metrics. For KITTI2012, we use Out-Noc and Out-All as metrics. They denote the percentage of erroneous pixels in non-occluded areas (Out-Noc) and total areas (Out-All). For KITTI2015, we use D1-bg, D1-fg, D1-all as metrics. They compute the percentage of stereo disparity outliers with errors greater than 3 pixels for the background (D1-bg), foreground (D1-fg), and all (D1-all) pixels, respectively. After fine-tuning for 50k iterations, we report the official results along with running time in Table~\ref{tab1}.
	
	As shown in the table, our method can make a trade-off between speed and accuracy. Although our MSCVNet is little less accurate than top-performance methods, such as GCNet\cite{kendall2017end}, PSMNet\cite{chang2018pyramid}, GANet\cite{zhang2019ga}. it is our advantage that we can achieve real-time at 41ms. Meanwhile, compared with the traditional method (SPS-St\cite{yamaguchi2014efficient}) and other fast networks, we achieve significant performance improvement. In particular, compared with Fast DS-CS\cite{yee2020fast} based on ADCensus and DispNetC\cite{mayer2016large} based on 1D Correlation, our MSCVNet obviously better than them in all evaluation metrics on KITTI2012, and only a little worse than Fast DS-CS in D1-fg on KITTI2015, demonstrating that integrating ADCensus and 1D Correlation is effective. We also visualize the results in Fig.~\ref{fig4} to further prove our method's effectiveness.
	
	\subsection{Ablation Study on KITTI2015}
	
	We test the impact of different settings on the performance from the following aspects. First, we compare the performance difference between Multi-scale cost volumes and a single cost volume. We only use 1D Correlation to construct the Multi-scale cost volumes and compare the performance with our method. Finally, we test the performance difference caused by the number of cascade hourglass networks and different loss functions.
	
	As shown in Table~\ref{tab2}, by comparing 1, 2 with 3, we can find that combining ADCensus and 1D Correlation can significantly improve accuracy while only increases 10ms. By comparing 3 with 4, we find that using two hourglass networks is better than one. This is obviously due to the increase of network layers. From the results of 4 and 5, it can be seen that using different methods to generate cost volume is significantly better than using only one method. We infer that this is due to redundancy in the network. Finally, by comparing 4 with 6, we can prove that calculating the loss for different areas can improve accuracy. All these results confirm that our network is efficient.
	
	\section{Conclusion}
	
	This paper proposes a fast and accurate stereo matching method for real-time robot applications. We integrate the traditional method and CNN to improve the quality of the 3D cost volume and construct a novel cascade hourglass network for cost aggregation.  We also design a novel algorithm for loss function by distinguishing discontinuous disparity areas. Results on Scene Flow and KITTI demonstrate the effectiveness of our MSCVNet. We can achieve a competitive result with a significantly fast speed at 41ms. We attempt to expand our work to multi-view stereo and combine depth completion to achieve a more accurate effect in future work.
	
	\section*{Acknowledgement}
	This work was supported by the National Key Research and Development Program of China (No. 2018YFB0204301), NSFC (No.61872374) and State Key Laboratory of High Performance Computing.

	\bibliographystyle{IEEEtran}
	\bibliography{IEEEabrv,refs}
	
\end{document}